%% file: arxiv.tex
\documentclass[conference]{IEEEtran}
\IEEEoverridecommandlockouts

\usepackage{cite}
\usepackage{amsmath,amssymb,amsfonts}
\usepackage{booktabs,multirow}
\usepackage{float}

\usepackage{algorithmic}
\usepackage{graphicx}
\usepackage{textcomp}
\usepackage{xcolor}
\usepackage{url}
\usepackage{hyperref}
\usepackage{subcaption}
\newif\ifappendix
 \appendixtrue  
\def\BibTeX{{\rm B\kern-.05em{\sc i\kern-.025em b}\kern-.08em
    T\kern-.1667em\lower.7ex\hbox{E}\kern-.125emX}}
\begin{document}

\title{Quantifying the Generation Modality Gap \\in Speech-Text Language Models}

\author{\IEEEauthorblockN{ Ju-Chieh Chou}
\IEEEauthorblockA{\textit{TTI-Chicago} \\
jcchou@ttic.edu}
\and
\IEEEauthorblockN{ Jiawei (Joe) Zhou}
\IEEEauthorblockA{
\textit{Stony Brook University}\\
jiawei.zhou.1@stonybrook.edu}
\and
\IEEEauthorblockN{ Karen Livescu}
\IEEEauthorblockA{
\textit{\textit{TTI-Chicago}}\\
klivescu@ttic.edu}

}

\maketitle

\begin{abstract}
Pure speech language models often lag behind text and speech-text language models in generating coherent content, but this gap is difficult to quantify because speech and text systems are typically evaluated with different metrics and trained on different data. We study the speech-text modality gap in a family of spoken language models, based on flow matching for continuous acoustic feature generation.  We construct a unified generation-based evaluation suite that compares speech-only, text-only, and speech-text language models trained on matched data distributions and evaluated in matched generation settings. We evaluate generated continuations along multiple dimensions: semantic coherence, measured by transcribing generated speech and scoring it with a reference language model; local phonetic structure, measured by phone n-gram distributional statistics; speaker consistency and acoustic quality; and emotion-based distributional metrics. Across datasets, we find that joint speech-text modeling substantially improves semantic coherence. However, the improvement is not uniform across metrics: phone-level metrics change only modestly, speaker similarity and predicted quality are lower for speech-text continuations, while emotion-based distributional metrics improve. Compared with larger-scale speech-only models, our speech-text model closes much of the scaling gap in transcript-based semantic coherence, providing evidence that text provides an efficient semantic training signal for spoken language modeling.


\end{abstract}


\section{Introduction}
\begin{figure}[h]
    \centering
    \includegraphics[width=0.83\linewidth]{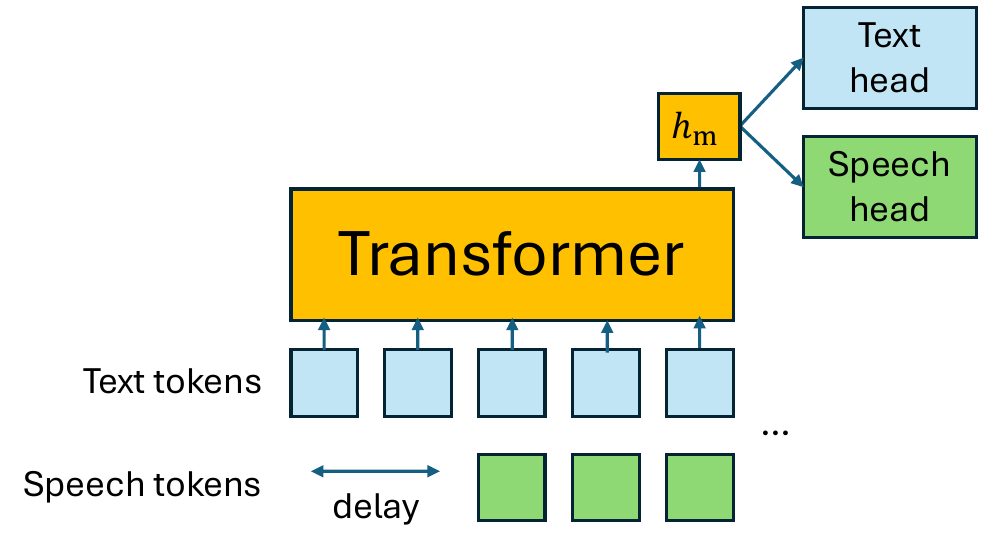}
    \caption{Speech-text LM architecture. The transformer processes text and speech streams and produces a context vector $h_m$ at the $m^\textrm{th}$ time step, which is used by the text and speech prediction heads to generate future text and speech output. The speech-only and text-only models use the same backbone with only the corresponding stream and output head. The speech head includes both a discrete token head (eq.~\ref{eq:sem}) and a conditional flow matching (CFM) head (eq.~\ref{eq:cfm}).}
    \label{fig:arch}
\end{figure}

Pre-training is the foundation of text language models (LMs). Recent work has explored the application of similar language modeling objectives to speech, producing spoken language models (SLMs)~\cite{lakhotia2021generative,chou2025flow,sugiura2025llama,arora2025landscape}. Unlike text, speech generation requires handling multiple factors at once: The generated signal must carry linguistic content, express that content as a plausible phone sequence, and include acoustic and paralinguistic properties such as speaker identity, emotion, and audio environment. Because speech generation requires modeling both content and acoustic realization, SLMs must devote modeling capacity to acoustic, speaker, and timing variation that text LMs do not need to model. Inner monologue speech-text LMs~\cite{defossez2024moshi,hu2025salm} aim to reduce this burden by incorporating text as an additional stream. This raises a quantitative question: How much (if any) linguistic capability is lost when moving from text to speech, and how much of this gap is recovered by adding text?  In addition, what are the tradeoffs in phonetic and acoustic properties when including or excluding a text stream?  


Answering these questions is not straightforward, because existing models are often trained on different data and evaluated differently. Results with existing speech-only and speech-text language models show that language modeling objectives can be used for high-quality speech generation, and that speech-text models can improve over speech-only models~\cite{nguyen2025spirit,maimon2025scaling,defossez2024moshi}. However, existing evaluations often focus on downstream performance~\cite{fang2025llama,ghosh2026audio,lu2024desta,lu2026desta2} or scaling behavior~\cite{cuervo2024scaling,maimon2025scaling,ramapuram2026scaling}, rather than isolating how text conditioning during pre-training affects different aspects of generated speech. This issue is especially important when analyzing pre-training: Evaluations after downstream adaptation or post-training may reflect task tuning rather than the effect of the pre-training objective itself, and introduces even more variability in training data and strategies. To understand the role of text in spoken language modeling, we need a matched training, data and evaluation protocol that compares speech-only, speech-text, and text-only models under the same generation setting and separately measures semantic, phonetic, and acoustic properties.

We design an evaluation protocol based on prompted generation and study a family of models with pure speech, speech-text, and text-only variants (Fig.~\ref{fig:arch}). Our model family is based on Flow-SLM~\cite{chou2025flow}, an SLM based on continuous speech representation generation, which we extend into a joint speech-text model to allow us to study all three variants. The text-only variant generates text continuations, while speech-only and speech-text models generate speech continuations. We evaluate these continuations with complementary metrics: transcript-based generation perplexity for semantic coherence, phone n-gram statistics for local phonetic structure, speaker similarity and predicted quality and emotion-based distributional metrics for acoustic realization. This setup allows us to quantify the improvement from speech-only to speech-text models, and to determine which aspects of the generated speech are affected.

Using this evaluation setup, we find that text conditioning primarily improves semantic coherence. Speech-text models achieve much lower generation perplexity than speech-only models, indicating stronger linguistic continuation ability. In contrast, local phone-statistic metrics change only modestly, suggesting that speech-only models already capture much of the short-range phonetic structure. Acoustic metrics show a different pattern: Speaker similarity and predicted quality favor speech-only generations, while emotion-based distributional metrics improve with text conditioning. Training dynamics further show that the internal text stream improves first, followed by the speech stream, suggesting that the speech-text model first learns to generate an internal text plan that the speech stream can then use for generation. Finally, we compare our models to a much larger-scale speech-only model, Llama-Mimi~\cite{sugiura2025llama}, and find that our text-conditioned SLMs can close much of the semantic gap to larger-scale speech-only models.

This work makes two main contributions. First, we introduce our matched training and evaluation suite for comparing speech-only, speech-text, and text-only language models under a shared generation setting, with semantic, phonetic, and acoustic evaluations. Second, we use this framework to characterize the effect of text conditioning during spoken language model pre-training in a family of SLMs, showing that text accelerates convergence and improves semantic coherence, while phonetic and acoustic metrics are affected in a less uniform way.  Our models, code, and evaluation setup will be made publicly available upon publication.

\section{Related work}
\label{sec:related}

\paragraph{Pure speech and speech-text language models}
Spoken language models extend language modeling objectives from text to speech, either through high-level (linguistically meaningful) discrete speech units~\cite{lakhotia2021generative}, multi-level (linguistic and acoustic) discrete codec representations~\cite{borsos2023audiolm,sugiura2025llama}, or continuous speech representations~\cite{chou2025flow,rouard2025continuous,yang2025generative}.
Speech-text language models are similar to these pure speech models, but also incorporate text either via interleaved speech-text  tokens~\cite{chou2023toward,nguyen2025spirit} or by adding an internal text stream (``inner monologue'')~\cite{defossez2024moshi}. 
Our work studies speech-only, (inner monologue) speech-text, and text-only models under matched training and evaluation.

In this work, we use the Flow-SLM~\cite{chou2025flow} architecture, a continuous representation model, as the speech generation backbone. Other recent continuous or multi-level speech codec
LM architectures could also be used for our purposes~\cite{rouard2025continuous,defossez2024moshi,sugiura2025llama}.  We choose Flow-SLM because it provides a good performance vs.~scale operating point, achieving similar semantic metric performance to higher-resource models while improving acoustic quality.  This allows us to study a range of experimental conditions on a limited compute budget (but we also include a comparison with a larger-scale model, Llama-Mimi~\cite{sugiura2025llama}, in some of our experiments).
Our goal is not to compare speech generation architectures, but to start with a strong speech backbone and keep it fixed while analyzing how adding text changes its generation behavior.


\paragraph{Modality gap analyses}
Several recent studies analyze the capability gap between speech and text models. These studies are the most closely related prior work, but they study different questions than we do. Hsu et al.~\cite{hsu2026anatomy} study speech-aware text LMs, where speech is used as input, but the model generates text. They study internal representations of speech-aware text LMs and their effect on downstream tasks. In contrast, our work studies pure speech LMs and speech-text LMs that can generate speech.

Wang et al.~\cite{wang2026speech} study speech generation by gradually changing the output modality of language models, from text to phones and then to tokens derived from HuBERT~\cite{hsu2021hubert} representations. They find that the transition from duplicated phones to HuBERT tokens results in the largest degradation in task performance. Our work instead focuses on models that use speech and text together, and asks how the two streams affect generation performance by controlling the training and evaluation conditions.



\section{Methods}
\label{sec:method}

This section describes our models (Fig.~\ref{fig:arch}) and general approach to studying the modality gap.  Section~\ref{sec:exp} provides more details about design choices and evaluation metrics.  



\subsection{Speech representation and loss}\label{sec:speech_model}

Our speech-only and speech-text models use the same speech representation and speech generation objective, modified from Flow-SLM~\cite{chou2025flow}.\footnote{Using the codebase at \url{https://github.com/jjery2243542/flow-slm}. We add dependence on discrete tokens (eq.~\ref{eq:sem}), and use x-prediction for the CFM head (eq.~\ref{eq:cfm}).} A speech tokenizer (here, Mimi~\cite{defossez2024moshi}) maps each waveform to two synchronized sequences: discrete (linguistically meaningful) tokens\footnote{These are often referred to as ``semantic tokens'' in the literature, but mainly carry phonetic information.} $z=(z_1,\ldots,z_M)$ with $z_m \in \mathcal{V_{\textrm{discrete}}}$
and continuous speech representations
$x=(x_1,\ldots,x_M)$ with $x_m \in \mathbb{R}^d$. 
The model predicts these two components with two heads (both contained within the ``speech head" of Fig.~\ref{fig:arch}): a discrete token head that predicts future discrete tokens, and a conditional flow matching (CFM)~\cite{lipman2022flow} head that predicts continuous speech representations.

At each time step $m$, a causal transformer produces a hidden state $h_m$ from the available context. In the speech-only model, this context contains only past speech; in the speech-text model, it also contains the available text-stream states (Fig.~\ref{fig:arch}). The discrete token head predicts the next $k$ discrete tokens using an autoregressive factorization (where $k=4$ following~\cite{chou2025flow}):
\begin{equation}
\label{eq:ar_fac}
p_\theta(z_{m:m+k-1}\mid h_m)=\prod_{i=0}^{k-1}
p_\theta(z_{m+i}\mid h_m,z_{m:m+i-1}),    
\end{equation}
where we condition on the previous semantic tokens by concatenating the token embeddings to $h_m$.  The loss for semantic tokens is
\begin{equation}
\label{eq:sem}
\mathcal{L}_{\mathrm{discrete}}=\mathbb{E}_{m}[-
\log p_\theta(z_{m:m+k-1}\mid h_m)].
\end{equation}


The CFM head predicts the continuous representation $x_m$ given the discrete token sequence $z_{m:m+k-1}$. 
We use an x-prediction parameterization with a velocity loss~\cite{li2026back},
\begin{equation}
\hat{x}_m=f_\theta(x_{m,\tau},\tau,h_m,z_{m:m+k-1}).
\end{equation}
For each target $x_m$, we sample Gaussian noise $\epsilon\sim\mathcal{N}(0,I)$ and time $\tau$ and construct
$x_{m,\tau}=(1-\tau)\epsilon+\tau x_m$.   
The CFM head predicts the clean target representation $x_m$.
We compute the CFM loss in velocity space: 
\begin{equation}
\label{eq:cfm}
\mathcal{L}_{\mathrm{CFM}}=\mathbb{E}_{m,\tau,\epsilon}[
\Vert
\hat{v}_{m,\tau}-v_{m}
\Vert_2^2]. 
\end{equation}
The target velocity for the linear path is
$v_{m}=x_m-\epsilon$,    
and the velocity implied by the clean prediction is $\hat{v}_{m,\tau}=(\hat{x}_{m}-x_{m,\tau})/({1-\tau})$. 
Finally, the full speech loss is
\begin{equation}
\label{eq:speech}
\mathcal{L}_{\mathrm{speech}}=
\mathcal{L}_{\mathrm{discrete}}
+
\mathcal{L}_{\mathrm{CFM}}.
\end{equation}

During inference, discrete tokens are sampled from the discrete token head. The CFM head then generates continuous speech representations conditioned on the hidden state and sampled discrete tokens, and the tokenizer (Mimi) decoder converts the generated representations to a waveform.

\subsection{Model classes}

We evaluate three model classes, and in all cases initialize the models from a pre-trained text LM. 

\paragraph{Text-only language model}
The text-only model is fine-tuned on transcriptions from a speech training set, using a standard next-token objective, which we refer to as $\mathcal{L}_{\mathrm{text}}$.
\paragraph{Speech-only language model}
The speech-only model generates speech from speech history alone, using the speech representation and loss defined in Sec.~\ref{sec:speech_model}. Its hidden state $h_m$ is computed only from past speech context, and its objective is
$\mathcal{L}_{\mathrm{speech}}$ (eq.~\ref{eq:speech}).

\paragraph{Speech-text language model}
The speech-text model extends the speech-only model with a text stream (inner monologue~\cite{defossez2024moshi}).  Here the speech hidden state $h_m$ is conditioned on the joint speech-text context rather than the speech history alone. 
The training objective combines the speech and text losses above:
$\mathcal{L}_{\mathrm{st}}=
\mathcal{L}_{\mathrm{text}}
+
\mathcal{L}_{\mathrm{speech}}$,
where the text loss is applied to the text head's prediction, and the speech loss to the speech prediction.
Thus, the speech-only and speech-text models use the same speech targets and speech losses; they differ in whether an internal text stream is available as additional context during speech generation.

\subsection{Speech-text modeling with delayed speech stream}

The speech-text model does not use word-level or frame-level speech-text alignment. Instead, we use a fixed text-leading schedule in both training and inference as in~\cite{hu2025salm,defossez2024moshi}. The text stream is shifted ahead of the speech stream by $D$ frames. During training, this means that when the model predicts speech at a given speech position, the speech stream can attend to the speech history and to text states that include a short text prefix ahead of the current speech position. 

At inference time, for prompted generation the model is given the speech prompt and its transcript. We first advance the text stream to be $D$ steps ahead of the speech stream (speech prompts are typically longer in terms of number of tokens). After this initial text, the model jointly generates the  text stream and the speech continuation. This schedule lets the text stream serve as an intermediate linguistic planning signal while avoiding explicit word-level or frame-level speech-text alignment as in~\cite{hu2025salm}.

\subsection{Evaluation approach}
We evaluate prompted and unprompted generations with semantic, phonetic, and acoustic metrics.
For prompted generation, a short spoken and/or written prompt is given, and the models generate sampled continuations. The metrics are generally computed on the generated continuation after prompt removal (with the exception of emotion-based metrics, which are based on emotion representations that require the full utterance), in order to evaluate the quality of only the generated content and not the prompt itself. 
For speech-only and speech-text models, we transcribe both the prompt and generated continuation using an automatic speech recognizer (ASR).  The ASR transcript may not preserve the prompt boundary exactly, so we remove the prompt portion via edit-distance matching.  Specifically, let $\tilde{c}_j$ be the full generated transcript and let $r_j$ be the ground-truth prompt transcript, both represented as character sequences. We estimate the prompt boundary as
\begin{equation}
e_j^\star
=
\arg\min_{0 \le e \le |\tilde{c}_j|}
d_{\mathrm{edit}}\!\left(\tilde{c}_{j,1:e}, r_j\right),
\end{equation}
where $d_{\mathrm{edit}}$ is character-level edit distance. The evaluated continuation is then the suffix
$\hat{c}_j
=
\tilde{c}_{j,e_j^\star+1:|\tilde{c}_j|}$.
We apply the same procedure to phone sequences for phone-level metrics. 
\paragraph{Semantic metric}
For semantic evaluation, we use generation perplexity (genPPL), which is commonly used to evaluate speech-only LMs~\cite{lakhotia2021generative,defossez2024moshi,rouard2025continuous},
as follows.  We convert all model outputs to text (using an ASR system for speech-only and speech-text models), score them with a reference language model, and compute the perplexity according to that model.  This provides a metric that can be 
compared across text-only, speech-only, and speech--text models. 
For concreteness, let $\{(r_j,\hat{c}_j)\}_{j=1}^{J}$ be the evaluated prompt-continuation pairs in text form. Here, $r_j$ is the prompt transcript.
For text-only models, $\hat{c}_j$ is the generated text continuation; for speech-only and speech--text models, $\hat{c}_j$ is the ASR transcript of the generated speech continuation after prompt removal. Let $\hat{c}_j=(\hat{c}_{j,1},\ldots,\hat{c}_{j,T_j})$ denote the continuation tokenized by the reference LM's tokenizer. We then have
\begin{equation}
\mathrm{genPPL}
=
\exp\left(
-
\frac{
\sum_{j=1}^{J}
\sum_{t=1}^{T_j}
\log p_{\mathrm{ref}}
\left(
\hat{c}_{j,t}
\mid
r_j, \hat{c}_{j,<t}
\right)
}{
\sum_{j=1}^{J} T_j
}
\right),
\end{equation}
where $p_{\mathrm{ref}}$ is the next-token probability computed by the reference LM. 
Lower genPPL indicates that the continuations are more likely under the reference LM given the prompts, and we use it as a proxy for semantic coherence.\footnote{Because genPPL uses a reference LM as a proxy,
it may be affected by domain mismatch between the reference LM and the evaluation data. To assess sensitivity to the choice of reference LM, we compute genPPL with several reference LMs and find highly consistent trends(Appendix~\ref{sec:diff_lm}). } 

\paragraph{Phonetic metric}
For speech-only and speech-text models, the generated speech is converted to phone sequences with a phone recognizer. We compute phone n-gram statistics over the generated continuations and compare them with the corresponding statistics from the reference continuations. We report Jensen-Shannon divergence (JSD) with respect to the reference distribution, denoted pJSD, as introduced in~\cite{ramapuram2026scaling}. Lower pJSD indicates that the generated speech better
matches the local phone statistics of the reference speech.

\paragraph{Acoustic metrics}
We evaluate acoustic properties using speaker similarity, predicted acoustic quality, and emotion distributional metrics.
We compute speaker similarity as the cosine similarity between the prompt and continuation using a speaker verification model.
We use predicted mean opinion score (MOS) as the acoustic quality. These metrics are commonly used to evaluate speech generation~\cite{le2023voicebox,chen2025f5,fang2025llama}. 
For emotion-based evaluation, we use an emotion classifier (emotion2vec+large~\cite{ma2024emotion2vec}) to obtain emotion label posteriors for generated and reference utterances, and compute the KL-divergence between them. We also compute Fréchet speech distance (FSD)~\cite{le2023voicebox} by encoding full generated and reference utterances with emotion2vec-base~\cite{ma2024emotion2vec} and measuring the Fréchet distance between their embedding distributions.

\section{Experiments}
\label{sec:exp}
\vspace{-.1in}
\subsection{Datasets}
To understand the impact of data distribution, we use three datasets:  Emilia~\cite{he2024emilia},  MLSEn~\cite{pratap2020mls}, and a synthesized version of C4~\cite{raffel2020exploring} (in-the-wild, audiobook, text-derived), for both training and evaluation. This allows us to study whether the speech-text modality gap is consistent across different training domains.
For C4, we use F5-TTS~\cite{chen2025f5} to synthesize sampled sentences ($\sim$2M). We sample at most one sentence from each C4 datapoint, using a reference speaker randomly sampled from LibriSpeech dev-clean~\cite{panayotov2015librispeech}. We filter out utterances with WER $>50\%$, resulting in $\sim$9k hours. To roughly match dataset sizes, we sample 10k-hour subsets from Emilia-En~\cite{he2024emilia} and from MLSEn~\cite{pratap2020mls}.

For evaluation, we create a set of prompts derived from $1500$ utterances, each of length $>6$s, from the test set corresponding to each of the three domains. For each utterance, we use the technique of~\cite{yeh2025whisper} to get a Whisper-based word-level alignment, find the word boundary closest to the first 3 seconds, and use the resulting $\sim$3s as the prompt. 

\subsection{Training}
We initialize all models from the OpenELM-270M pretrained text LM~\cite{hassid2023textually,mehta2024openelm}, and use the speech head architecture from~\cite{chou2025flow}. We use the pre-quantized representation from Mimi~\cite{defossez2024moshi} as the continuous representation $x$, and the first level of residual vector quantization (RVQ) as the discrete tokens $z$. For speech-text LMs, we use a delay $D=2$ frames. 
We train each model for 40k iterations (based on preliminary experiments), which corresponds to roughly two epochs over each dataset. We use a batch size of 128 utterances for MLSEn-10k and 160 utterances for Emilia and F5-C4, adjusted to account for different average utterance lengths across datasets and GPU memory constraints. We use a learning rate of $5 \times 10^{-4}$. 

\subsection{Evaluation details}
We sample 5 continuations per prompt, and sample $1500 \times 5$ utterances for unprompted generation. We apply simple model-specific decoding adjustments before evaluation. For speech-text LMs, the text stream may terminate while the speech stream continues generating instead of emitting an end-of-sequence token. To encourage synchronized termination, we introduce two decoding hyperparameters, $eos\_topk$ and $eos\_bias$: When the speech EOS token appears among the top-$eos\_topk$ candidates, we add $eos\_bias$ to its logit before sampling. We use $eos\_topk=150$ and $eos\_bias=15$. For speech-only LMs, generated sequences sometimes contain excessive silence. We identify silence-associated tokens using a VAD model\footnote{We use the VAD from the TorchAudio package~\cite{hwang2023torchaudio}.} and apply a logit penalty of $5$ to these tokens during sampling. These adjustments are used to reduce decoding artifacts that would otherwise dominate the evaluation. All other inference parameters follow Chou et al.~\cite{chou2025flow}.


After removing the prompt, some generated continuations are empty (e.g., when the 3-second prompt already forms a complete sentence), and we exclude these from evaluation. 
Specifically, for each metric, we evaluate on the joint non-empty subset: the set of examples for which all compared models produce a non-empty continuation for that metric.\footnote{For each model, $\sim$94\% of prompts produce non-empty continuations. After taking the intersection across models,
$\sim$80\% of examples remain.}
 
\subsubsection{Semantic metric}
To compute genPPL, we transcribe the generated continuations using Whisper small.en~\cite{radford2023robust} and score the resulting continuation transcripts with OLMo-3 7B~\cite{team2025olmo}.



\subsubsection{Phonetic metric}
We use  5-gram pJSD~\cite{ramapuram2026scaling}. For each generated speech continuation, we obtain a corresponding phone sequence using an audio-guided G2P model (POWSM-CTC)~\cite{li2025powsm}. Given the generated speech and its transcript from Whisper small.en, POWSM-CTC predicts the phone sequence. 

\begin{figure*}[t]
    \includegraphics[width=\linewidth]{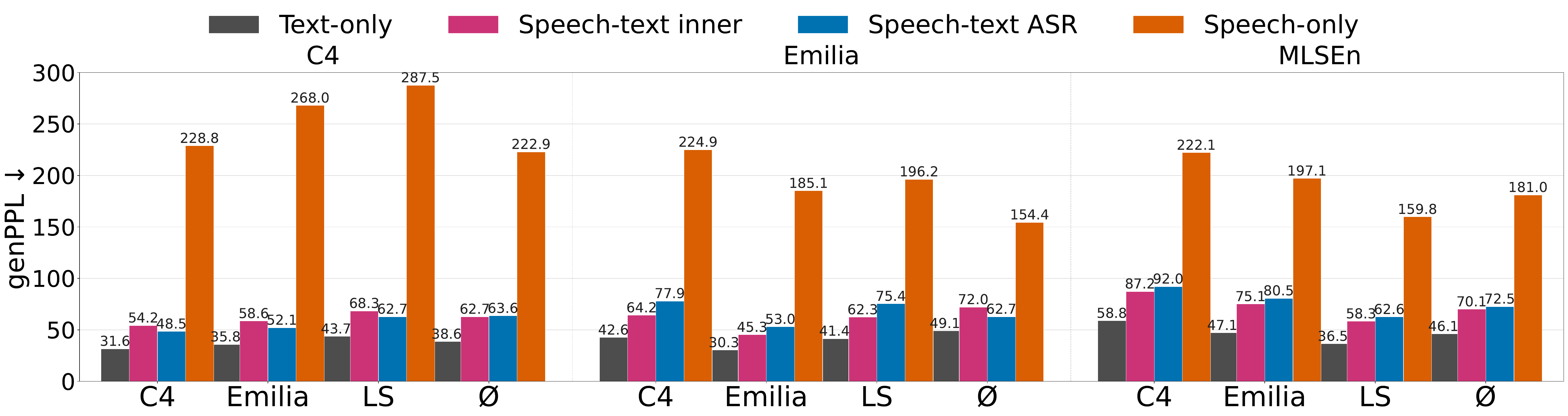}
    \caption{GenPPL across training sets and prompts. 
Each panel corresponds to a training set. X-axis labels indicate the prompt source;  $\emptyset$ is the unprompted setting.
We compare text-only generation, the internal text stream of the speech-text model (speech-text inner), ASR transcripts of speech-text continuations (speech-text ASR), and ASR transcripts of speech-only continuations. }
    \label{fig:ppl}
\end{figure*}

\subsubsection{Acoustic metrics}
 We use WavLM-large-TDNN, following~\cite{he2024emilia,le2023voicebox}\footnote{\url{https://github.com/microsoft/UniSpeech/tree/main/downstreams/speaker_verification}} as our speaker verification model. To reduce the effect of vocoder mismatch, we compare against a re-synthesized prompt rather than the original prompt waveform. 
We evaluate audio quality using UTMOS~\cite{saeki2022utmos}, which predicts the mean opinion score (MOS) of each generated continuation. We provide the detailed results in Appendix~\ref{sec:detailed}.


\section{Results}

\subsection{Internal text substantially reduces the semantic gap}



\begin{figure*}[t]
\centering

\begin{minipage}[t]{0.69\linewidth}
\centering
    \includegraphics[width=\linewidth]{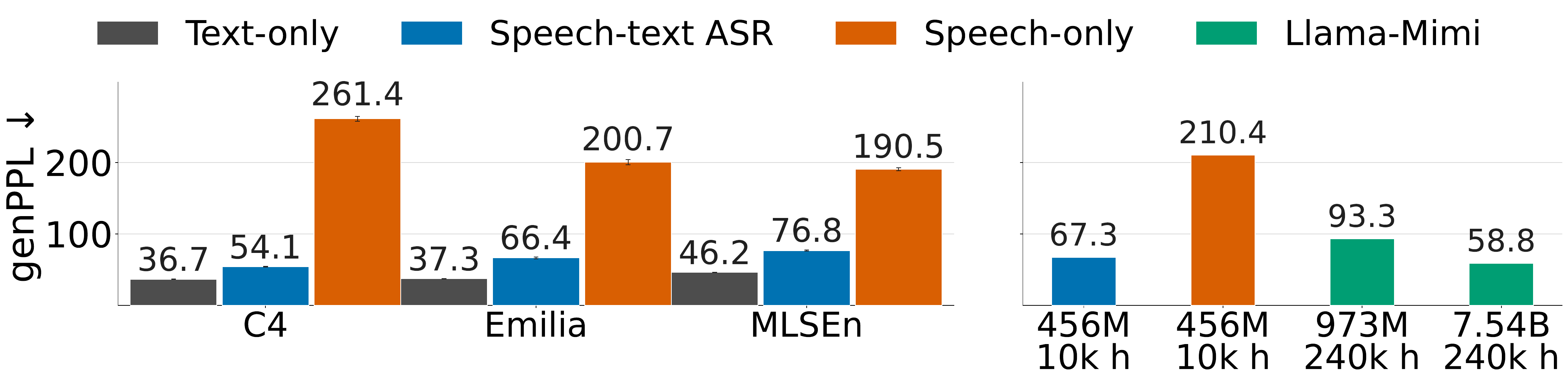}
\end{minipage}
\hfill
\begin{minipage}[t]{0.3\linewidth}
\centering
\includegraphics[width=\linewidth]{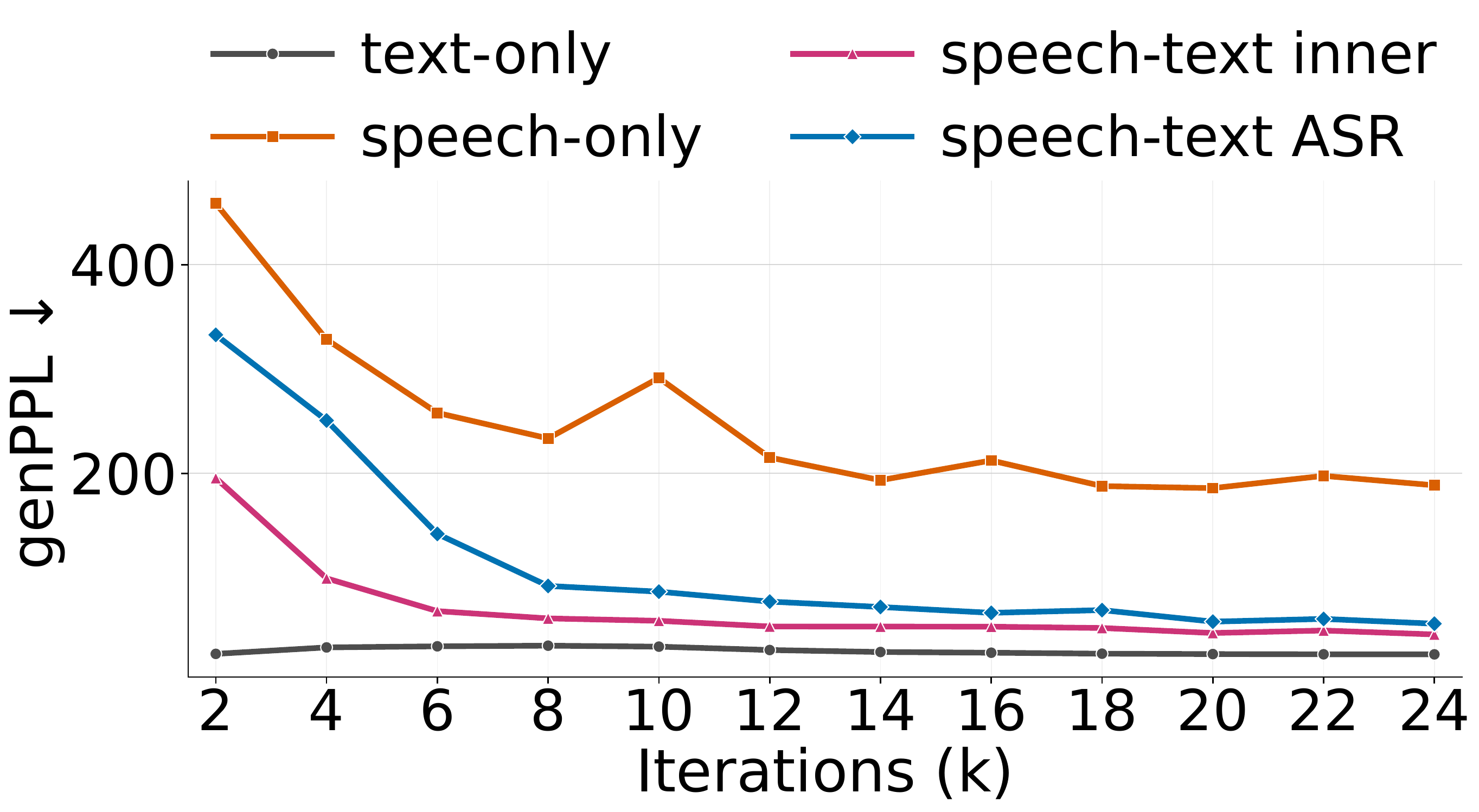}
\end{minipage}

\caption{
\textbf{Left:} GenPPL by training dataset, aggregated over all prompt sources and the unprompted setting. Error bars are 95\% confidence intervals estimated by bootstrap resampling.
\textbf{Middle:} GenPPL, aggregated over the training sets, vs.~Llama-Mimi (1.3B and 8B). The x-axis labels are the \#params excluding embedding layers (as in~\cite{kaplan2020scaling}) and the training set size.
\textbf{Right:} Training dynamics of genPPL when training on Emilia 10k.
}
\label{fig:gap_analysis}

\end{figure*}

Across training and evaluation prompt domains (Fig.~\ref{fig:ppl}), speech-only models have much higher genPPL than text-only and speech-text models, suggesting weaker linguistic competence.
Adding text closes much of this gap: Speech-text ASR genPPL is consistently far lower than speech-only genPPL. This also holds in the unprompted setting, where the text stream is self-generated,
suggesting that the internal text stream provides an effective linguistic planning signal.

Fig.~\ref{fig:gap_analysis}-left summarizes the average genPPL across evaluation domains. C4 has the largest gap between speech-only and speech-text ASR, while speech-text ASR remains closest to text-only. One possible explanation is that C4 contains more open-domain written-text content, including 
named entities that are difficult to learn from speech alone. 

\subsection{Local phone statistics change only modestly}
\begin{figure}[h]
    \centering
    \includegraphics[width=\linewidth]{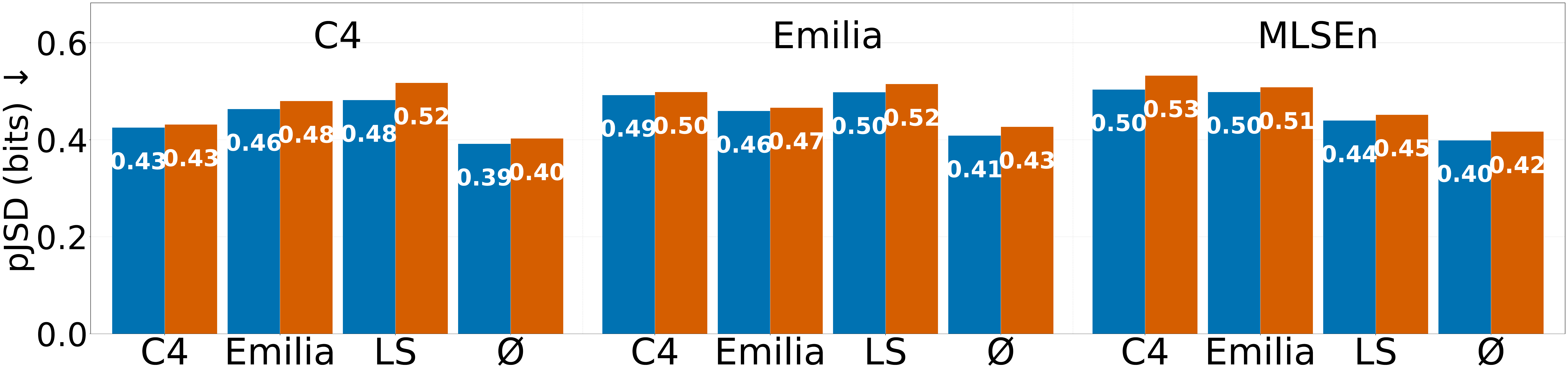}
    \caption{5-gram pJSD results. For unprompted generation ($\emptyset$), we compare to the prompt set that matches the training set. The legend follows Fig.\ref{fig:gap_analysis}}
    \label{fig:pjsd}
\end{figure}
Fig.~\ref{fig:pjsd} shows phone n-gram JSD between generated and reference continuations. Compared with the large genPPL differences in Fig.~\ref{fig:ppl}, the gap between speech-only and speech-text generation is small. Speech-text ASR continuations are only slightly closer to the reference phone n-gram distribution than speech-only continuations, suggesting that speech-only models already learn much of the local phonetic structure. 

\subsection{Acoustic metrics move in different directions}
\begin{figure}[h]

    \centering
    \includegraphics[width=\linewidth]{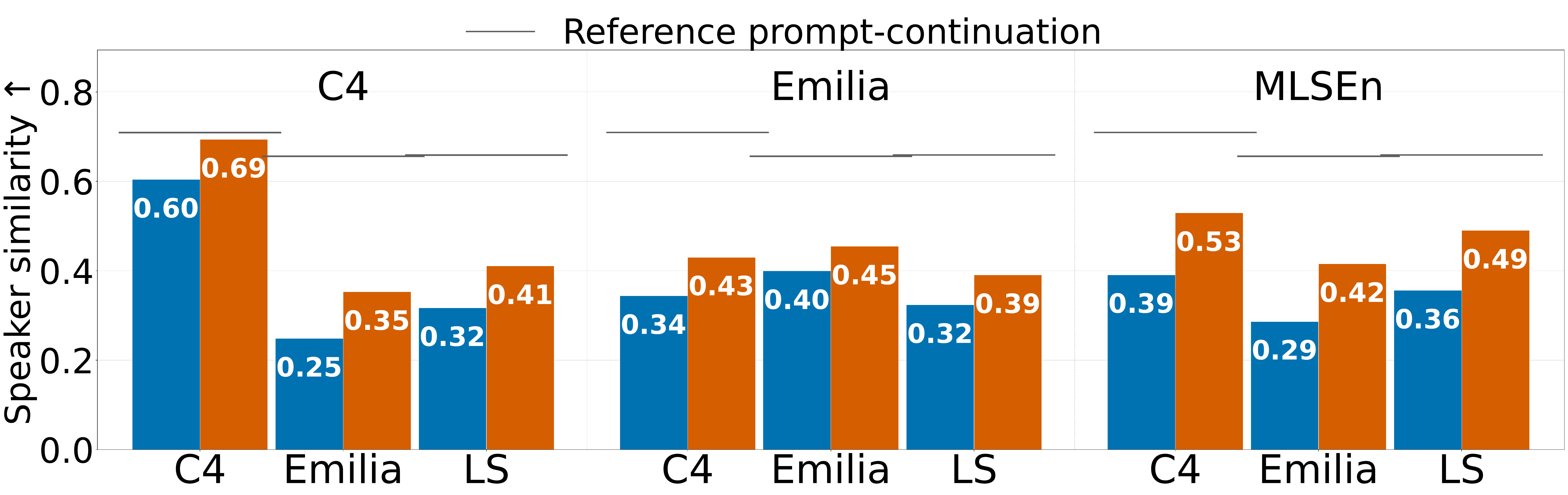}
    \caption{Speaker similarity from the speaker verification model. The legend follows Fig.\ref{fig:gap_analysis}.}
    \label{fig:speaker}
\end{figure}
\begin{figure}[h]
    \centering
    \includegraphics[width=\linewidth]{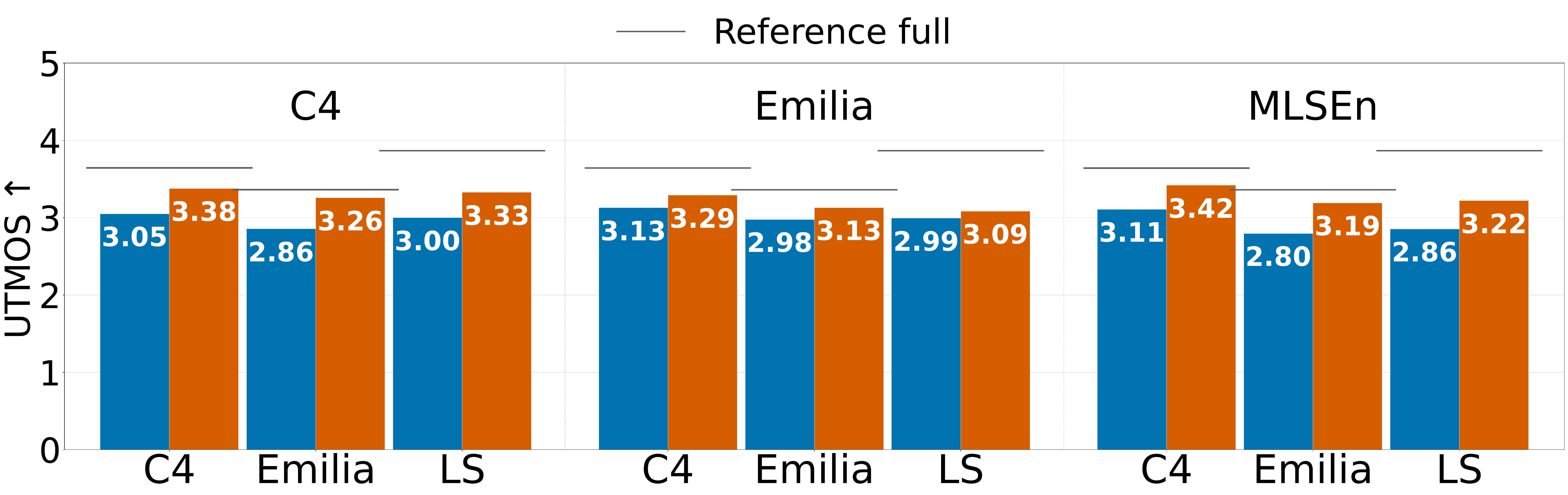}
    \caption{Predicted MOS from the UTMOS model~\cite{saeki2022utmos}. The legend follows Fig.\ref{fig:gap_analysis}.}
    \label{fig:mos}
\end{figure}
Fig.~\ref{fig:speaker} shows speaker similarity between the generated continuation and the prompt, and Fig.~\ref{fig:mos} shows predicted MOS from UTMOS. Both metrics favor speech-only over speech-text models. Therefore, the semantic gains from text conditioning do not necessarily translate into better acoustic quality or speaker preservation.
These results may also reflect a tradeoff in capacity allocation. Unlike the speech-only model, the speech-text model also optimizes a text-stream loss, potentially reducing the effective emphasis on acoustic properties. 

\begin{figure}
    \centering
    \includegraphics[width=\linewidth]{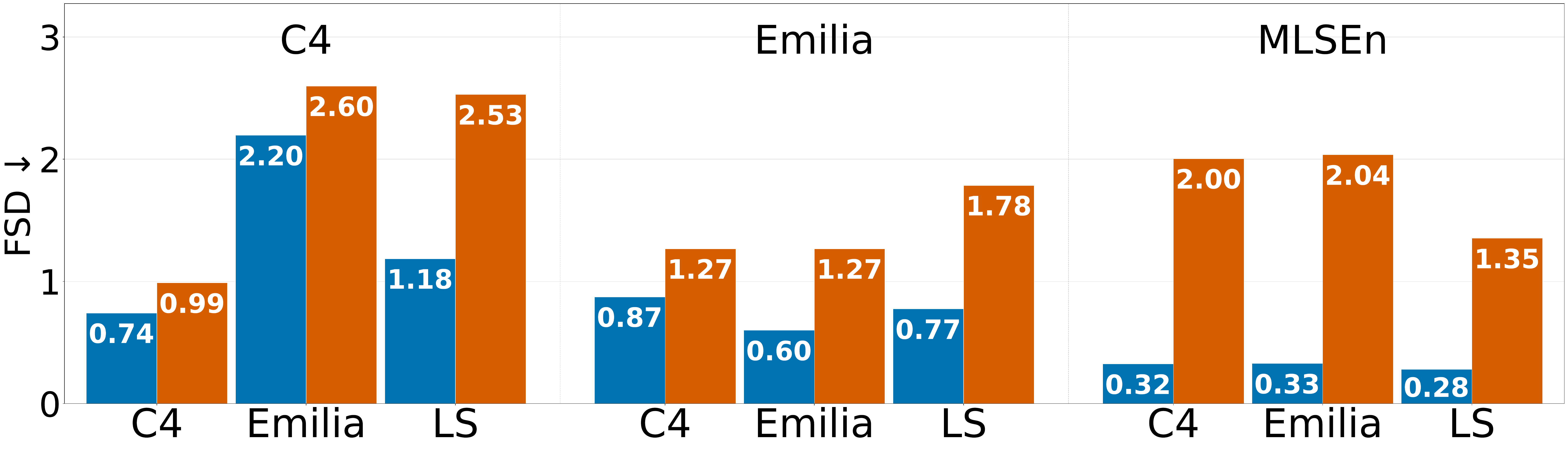}
    \caption{FSD using utterance-level embeddings from emotion2vec-base~\cite{ma2024emotion2vec}. The legend follows Fig.\ref{fig:gap_analysis}.}
    
    \label{fig:emo_fsd}
\end{figure}
\begin{figure}[!htbp]
    \centering
    \includegraphics[width=\linewidth]{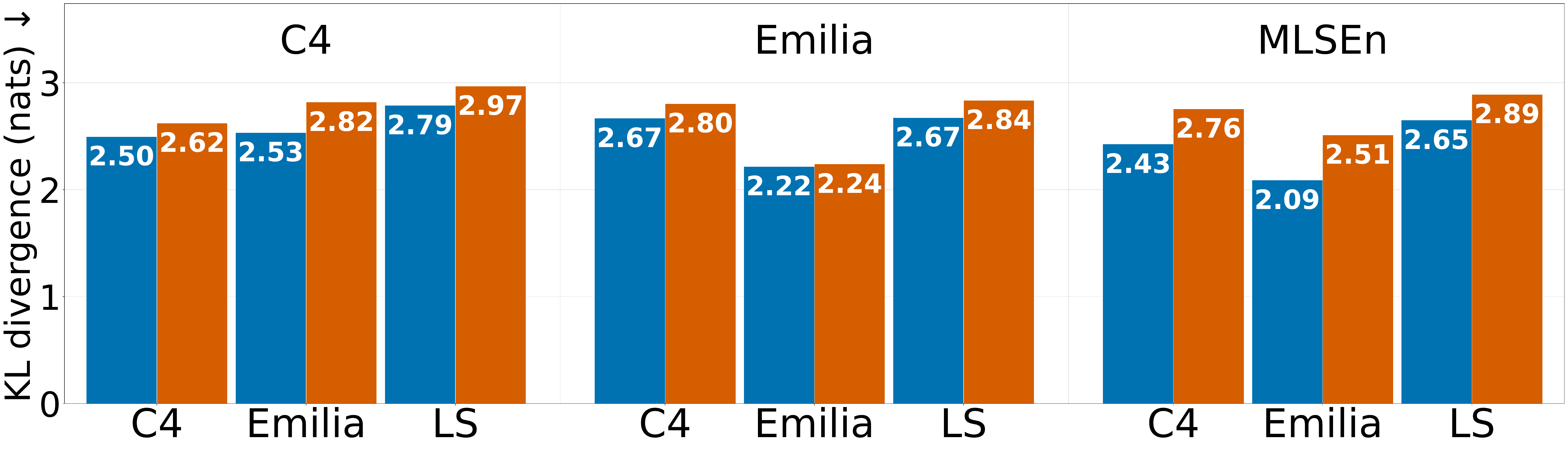}
    \caption{KL divergence between emotion2vec-large emotion-label posteriors for full reference and generated utterances. The legend follows Fig.\ref{fig:gap_analysis}.}
    \label{fig:emo_kl}
\end{figure}

For FSD computed using emotion2vec-base embeddings (Fig.~\ref{fig:emo_fsd}), speech-text LMs achieve lower distances to the reference distribution than speech-only LMs. The KL divergence between the emotion-label distributions of generated and reference utterances shows a similar trend (Fig. \ref{fig:emo_kl}). However, emotion2vec representations may encode both linguistic and acoustic information, and embedding similarity does not necessarily isolate emotion alone, as noted in~\cite{tsai2026false}. We therefore interpret these metrics as measuring overall emotion-related consistency rather than purely acoustic emotion preservation. 
\subsection{Text conditioning accelerates convergence}
To analyze training dynamics, we retrain the models for 24k iterations on Emilia 10k hours, corresponding to roughly one epoch, and plot genPPL throughout training (Fig.~\ref{fig:gap_analysis}-right). 

For the speech-text model, the internal text stream reaches low genPPL earlier than the speech stream, which improves more gradually but eventually catches up to the text stream.
This pattern is consistent with the model initialization: The text stream starts from a pretrained text LM and therefore adapts quickly to text continuation, while the speech pathway must learn to map this linguistic structure into recognizable speech. 
In contrast, the speech-only model improves more slowly and plateaus at a substantially higher genPPL.

\subsection{Comparison with Llama-Mimi}

Finally, we compare our models with Llama-Mimi~\cite{sugiura2025llama} 1.3B and 8B (Fig.~\ref{fig:gap_analysis}-middle ).\footnote{Llama-Mimi models generate $\sim$$80\%$ non-empty continuations, and the joint non-empty subset contains $\sim$$60\%$ of the continuations. We use the inference parameters from \url{https://huggingface.co/llm-jp/Llama-Mimi-1.3B}.} Llama-Mimi fine-tunes a text Llama LM on flattened Mimi tokens from the first four RVQ levels,
using substantially more speech data ($240$k hours) and larger model sizes than we do. Under our genPPL evaluation, the Llama-Mimi models outperform our speech-only model, consistent with the benefits of larger-scale speech-only training. However, adding a text stream substantially closes this gap: Our speech-text model achieves much lower genPPL than our speech-only model and approaches the performance of the larger Llama-Mimi models. This suggests that text conditioning can compensate for a substantial part of the data and model-scale gap in terms of semantic continuation ability. Additional scaling results and a matched Llama-Mimi reproduction are provided in Appendix~\ref{sec:scaling}.

\section{Conclusion}
We introduce a matched training and evaluation suite to compare across text-only, speech-text, and speech-only language models. 
Our results show that inner text substantially reduces the semantic modality gap: Speech-text models achieve much lower genPPL than speech-only models. However, this improvement is not uniform across all metrics. Local phone n-gram statistics change only modestly, suggesting that speech-only models already capture much of the phonetic structure. Also, speech-only models achieve better speaker similarity and predicted quality, while speech-text models perform better on emotion-related distributional metrics. 

In this work, we study one family of models of the same size, and using the same scale of training data. We have seen that, when comparing to a higher-resource speech-only model, our speech-text model can recover much of the performance difference.  An interesting direction for future work would be to extend our analyses to different model classes, and to combine studies with scaling analyses to better understand scaling behavior across modalities.

\section{Acknowledgment}
We thank Shinji Watanabe for helpful discussions and feedback. The authors used ChatGPT to polish the text style, and OpenAI Codex as a coding assistant. 

\bibliographystyle{IEEEtran}
\bibliography{ref}

\clearpage

\ifappendix

\clearpage
\onecolumn
\appendix
\subsection{Detailed metrics}\label{sec:detailed}
We show the detailed metrics for each model and each prompt set in Table~\ref{tab:detail_vertical}. 
\input{all_vertical}
\clearpage
\subsection{Scaling results and reproduced Llama-Mimi}\label{sec:scaling}
We scale our flow-based model from 10k hours to 45k hours (iterations from 40k to 100k) and from the 270M backbone (corresponding to 456M parameters) to the 1.1B backbone (corresponding to 1.16B parameters) of OpenELM~\cite{mehta2024openelm}. To isolate the architecture change, we also reproduce Llama-Mimi under the matched setup with the base model. The results are shown in Table~\ref{tab:scaling}. We use Emilia training and test data. The speech-only model shows improvement in genPPL, while the speech-text model is more saturated. Our reproduced Llama-Mimi performs worse than our flow-based model. This could be either due to missing details in our reproduction or because it is less training-efficient, as it flattens the RVQ tokens.
\input{all_table_v2}
\clearpage
\subsection{Different reference LMs}\label{sec:diff_lm}
We compare different reference LMs around the 7B scale, including OLMo 7B~\cite{team2025olmo}, Ministral 8B~\cite{liu2026ministral}, Gemma 7B~\cite{team2024gemma}, Qwen 2.5~\cite{qwen2024qwen2}, and Llama 3.1 8B~\cite{grattafiori2024llama}.\footnote{Model checkpoints on Hugging Face: allenai/Olmo-3-1025-7B, mistralai/Ministral-3-8B-Base-2512, google/gemma-7b, Qwen/Qwen2.5-7B, meta-llama/Llama-3.1-8B.} The results are shown in Table~\ref{tab:ref_lm}. Spearman's rank correlation coefficients ($\rho$) between models are shown in Table~\ref{tab:spearman}. Generally the the reference LMs show the same trend, with $\rho > 0.98$ for each pair. 
\input{ref_lms}
\input{spearman_table}
\fi
\end{document}

%% file: all_vertical.tex
\begin{table}[H]
\centering
\caption{Detailed metrics for each model across training datasets and prompt sets. Parameter counts exclude the input embedding layer. Non-empty prompts for each datasets: C4 (4707 / 7500), Emilia (4733 / 7500) and LibriSpeech (4710 / 7500).}
\label{tab:detail_vertical}

\setlength{\tabcolsep}{4pt}
\renewcommand{\arraystretch}{1.04}
\resizebox{\textwidth}{!}{%
\begin{tabular}{@{}lllllrrrr@{}}
\toprule
Model & Training data & Variant & Params. & Prompts & genPPL $\downarrow$ & pJSD $\downarrow$ & SpkSim $\uparrow$ & UTMOS $\uparrow$ \\
\midrule
\midrule
\multirow{18}{*}{Flow-based} & \multirow{6}{*}{C4} & \multirow{3}{*}{Speech-only} & \multirow{3}{*}{456M} & C4  & 228.8 & 0.47 & 0.70 & 3.41 \\
 &  &  &  & Emilia  & 265.1 & 0.52 & 0.36 & 3.28 \\
 &  &  &  & LibriSpeech  & 291.6 & 0.56 & 0.42 & 3.35 \\
\cmidrule(lr){3-9}
 &  & \multirow{3}{*}{Speech-text} & \multirow{3}{*}{456M} & C4  & 49.9 & 0.47 & 0.62 & 3.09 \\
 &  &  &  & Emilia  & 51.8 & 0.51 & 0.26 & 2.90 \\
 &  &  &  & LibriSpeech  & 63.4 & 0.53 & 0.32 & 3.02 \\
\cmidrule(lr){2-9}
 & \multirow{6}{*}{Emilia10k} & \multirow{3}{*}{Speech-only} & \multirow{3}{*}{456M} & C4  & 223.5 & 0.54 & 0.44 & 3.34 \\
 &  &  &  & Emilia  & 183.3 & 0.51 & 0.47 & 3.19 \\
 &  &  &  & LibriSpeech  & 195.8 & 0.56 & 0.40 & 3.12 \\
\cmidrule(lr){3-9}
 &  & \multirow{3}{*}{Speech-text} & \multirow{3}{*}{456M} & C4  & 81.8 & 0.53 & 0.36 & 3.23 \\
 &  &  &  & Emilia  & 52.9 & 0.50 & 0.41 & 3.07 \\
 &  &  &  & LibriSpeech  & 78.7 & 0.55 & 0.33 & 3.04 \\
\cmidrule(lr){2-9}
 & \multirow{6}{*}{MLSEn10k} & \multirow{3}{*}{Speech-only} & \multirow{3}{*}{456M} & C4  & 220.7 & 0.57 & 0.53 & 3.43 \\
 &  &  &  & Emilia  & 198.0 & 0.55 & 0.42 & 3.22 \\
 &  &  &  & LibriSpeech  & 160.0 & 0.50 & 0.49 & 3.23 \\
\cmidrule(lr){3-9}
 &  & \multirow{3}{*}{Speech-text} & \multirow{3}{*}{456M} & C4  & 92.4 & 0.54 & 0.39 & 3.12 \\
 &  &  &  & Emilia  & 80.9 & 0.54 & 0.29 & 2.82 \\
 &  &  &  & LibriSpeech  & 63.4 & 0.49 & 0.36 & 2.86 \\
\midrule
\multirow{9}{*}{Llama-Mimi} & \multirow{3}{*}{Emilia10k} & \multirow{3}{*}{Matched reproduced} & \multirow{3}{*}{241M} & C4  & 407.9 & 0.52 & 0.40 & 3.10 \\
 &  &  &  & Emilia  & 283.0 & 0.50 & 0.45 & 3.14 \\
 &  &  &  & LibriSpeech  & 328.3 & 0.55 & 0.38 & 2.94 \\
\cmidrule(lr){2-9}
 & \multirow{6}{*}{240k h speech} & \multirow{6}{*}{Published} & \multirow{3}{*}{973M} & C4  & 106.8 & 0.52 & 0.48 & 3.07 \\
 &  &  &  & Emilia  & 86.5 & 0.50 & 0.46 & 2.93 \\
 &  &  &  & LibriSpeech  & 80.0 & 0.47 & 0.44 & 2.92 \\
\cmidrule(lr){4-9}
 &  &  & \multirow{3}{*}{7.54B} & C4  & 68.3 & 0.51 & 0.49 & 3.03 \\
 &  &  &  & Emilia  & 53.0 & 0.50 & 0.48 & 2.90 \\
 &  &  &  & LibriSpeech  & 48.7 & 0.47 & 0.45 & 2.92 \\
\bottomrule
\end{tabular}
}
\end{table}

%% file: all_table_v2.tex
\begin{table}[H]
\centering
\caption{Scaling results on Emilia, including data scaling, model scaling, and a matched Llama-Mimi reproduction to isolate architectural differences. Metrics are computed on the common subset of 4,644/7,500 non-empty continuations. Parameter counts exclude the input embedding layer.}
\label{tab:scaling}
\setlength{\tabcolsep}{5pt}
\renewcommand{\arraystretch}{1.08}
\resizebox{\textwidth}{!}{%
\begin{tabular}{@{}lllcrrrr@{}}
\toprule
Model & Training data & Variant & Params. & genPPL $\downarrow$ & pJSD $\downarrow$ & SpkSim $\uparrow$ & UTMOS $\uparrow$ \\
\midrule
\midrule
\multirow{2}{*}{Base} & \multirow{2}{*}{Emilia10k} & Speech-only & 456M & 182.1 & 0.51 & 0.47 & 3.19 \\
 &  & Speech-text & 456M & 52.7 & 0.51 & 0.41 & 3.07 \\
\midrule
\multirow{2}{*}{Base + data scale} & \multirow{2}{*}{Emilia45k} & Speech-only & 456M & 149.3 & 0.51 & 0.50 & 3.25 \\
 &  & Speech-text & 456M & 51.6 & 0.50 & 0.45 & 3.22 \\
\midrule
\multirow{2}{*}{Base + model scale} & \multirow{2}{*}{Emilia10k} & Speech-only & 1.16B & 170.2 & 0.51 & 0.48 & 3.19 \\
 &  & Speech-text & 1.16B & 54.5 & 0.51 & 0.41 & 3.05 \\
\midrule
\multirow{3}{*}{Llama-Mimi} & Emilia10k & Matched reproduced & 241M & 282.6 & 0.50 & 0.45 & 3.14 \\
 & \multirow{2}{*}{240k h speech} & Published & 973M & 86.5 & 0.50 & 0.45 & 2.94 \\
 &  & Published & 7.54B & 52.6 & 0.50 & 0.47 & 2.90 \\
\bottomrule
\end{tabular}
}
\end{table}

%% file: ref_lms.tex
\begin{table}[H]
\centering
\caption{GenPPL computed using five different reference LMs, across all model and evaluation settings.}
\label{tab:ref_lm}
\resizebox{\textwidth}{!}{%
\begin{tabular}{lllrrrrr}
\toprule
Prompt set & Dataset & Mode & OLMo 7B & Ministral 8B & Gemma 7B & Qwen2.5 7B & Llama 3.1 8B \\
\midrule
C4 & C4 & Speech--text ASR & 48.3 & 40.5 & 42.9 & 43.1 & 45.0 \\
C4 & C4 & Speech--text inner & 54.0 & 42.4 & 46.2 & 47.2 & 48.6 \\
C4 & C4 & Text-only & 31.4 & 25.2 & 26.3 & 26.8 & 27.3 \\
C4 & C4 & Speech-only & 230.4 & 202.7 & 216.8 & 221.7 & 223.9 \\
C4 & Emilia10k & Speech--text ASR & 78.4 & 77.7 & 82.5 & 85.0 & 90.7 \\
C4 & Emilia10k & Speech--text inner & 64.4 & 62.3 & 65.4 & 68.4 & 73.8 \\
C4 & Emilia10k & Text-only & 42.5 & 38.5 & 40.3 & 41.4 & 45.3 \\
C4 & Emilia10k & Speech-only & 227.6 & 233.1 & 241.5 & 249.1 & 255.3 \\
C4 & MLSEn10k & Speech--text ASR & 92.1 & 84.8 & 95.4 & 90.1 & 92.9 \\
C4 & MLSEn10k & Speech--text inner & 87.2 & 79.3 & 89.4 & 84.2 & 84.8 \\
C4 & MLSEn10k & Text-only & 58.7 & 53.3 & 60.5 & 56.3 & 57.0 \\
C4 & MLSEn10k & Speech-only & 223.8 & 211.9 & 250.2 & 229.3 & 228.8 \\
\midrule
Emilia & C4 & Speech--text ASR & 52.1 & 46.0 & 46.6 & 50.9 & 49.0 \\
Emilia & C4 & Speech--text inner & 58.6 & 48.5 & 50.8 & 55.7 & 53.7 \\
Emilia & C4 & Text-only & 35.6 & 30.5 & 30.6 & 34.1 & 32.6 \\
Emilia & C4 & Speech-only & 269.9 & 215.7 & 220.8 & 243.6 & 238.8 \\
Emilia & Emilia10k & Speech--text ASR & 53.0 & 50.0 & 47.7 & 53.2 & 51.2 \\
Emilia & Emilia10k & Speech--text inner & 45.1 & 41.7 & 39.7 & 44.6 & 43.3 \\
Emilia & Emilia10k & Text-only & 30.2 & 28.0 & 26.8 & 29.3 & 28.9 \\
Emilia & Emilia10k & Speech-only & 186.5 & 170.7 & 164.6 & 184.6 & 175.6 \\
Emilia & MLSEn10k & Speech--text ASR & 80.5 & 71.4 & 74.3 & 76.9 & 73.8 \\
Emilia & MLSEn10k & Speech--text inner & 75.0 & 64.9 & 68.9 & 69.8 & 66.5 \\
Emilia & MLSEn10k & Text-only & 46.9 & 41.2 & 43.3 & 43.6 & 41.7 \\
Emilia & MLSEn10k & Speech-only & 198.3 & 176.3 & 193.2 & 193.5 & 179.9 \\
\midrule
LibriSpeech & C4 & Speech--text ASR & 62.7 & 55.1 & 57.2 & 61.9 & 58.7 \\
LibriSpeech & C4 & Speech--text inner & 68.3 & 56.2 & 59.6 & 65.1 & 62.3 \\
LibriSpeech & C4 & Text-only & 43.6 & 36.7 & 37.7 & 41.6 & 39.8 \\
LibriSpeech & C4 & Speech-only & 289.9 & 242.1 & 249.5 & 275.3 & 259.0 \\
LibriSpeech & Emilia10k & Speech--text ASR & 75.6 & 70.7 & 72.6 & 79.7 & 73.6 \\
LibriSpeech & Emilia10k & Speech--text inner & 62.4 & 56.8 & 57.4 & 64.2 & 60.0 \\
LibriSpeech & Emilia10k & Text-only & 41.3 & 36.3 & 37.7 & 40.0 & 38.9 \\
LibriSpeech & Emilia10k & Speech-only & 198.4 & 185.4 & 186.1 & 208.9 & 187.9 \\
LibriSpeech & MLSEn10k & Speech--text ASR & 62.3 & 52.1 & 57.0 & 56.6 & 54.7 \\
LibriSpeech & MLSEn10k & Speech--text inner & 58.0 & 47.7 & 52.4 & 51.6 & 49.7 \\
LibriSpeech & MLSEn10k & Text-only & 36.2 & 29.1 & 32.1 & 31.0 & 30.0 \\
LibriSpeech & MLSEn10k & Speech-only & 160.9 & 140.0 & 158.0 & 158.3 & 144.4 \\
\bottomrule
\end{tabular}%
}
\end{table}

%% file: spearman_table.tex
\begin{table}[H]
\centering
\caption{Pairwise Spearman rank correlations between reference LMs over the same 36 genPPL values.}
\label{tab:spearman}
\small
\begin{tabular}{lrrrrr}
\toprule
Oracle LM & OLMo 7B & Ministral 8B & Gemma 7B & Qwen2.5 7B & Llama 3.1 8B \\
\midrule
OLMo 7B      & 1.00 & 0.99 & 0.99 & 0.99 & 0.99 \\
Ministral 8B & 0.99 & 1.00 & 0.99 & 1.00 & 1.00 \\
Gemma 7B     & 0.99 & 0.99 & 1.00 & 0.99 & 0.99 \\
Qwen2.5 7B   & 0.99 & 1.00 & 0.99 & 1.00 & 1.00 \\
Llama 3.1 8B & 0.99 & 1.00 & 0.99 & 1.00 & 1.00 \\
\bottomrule
\end{tabular}
\end{table}

%% file: arxiv.bbl
\begin{thebibliography}{10}
\providecommand{\url}[1]{#1}
\csname url@samestyle\endcsname
\providecommand{\newblock}{\relax}
\providecommand{\bibinfo}[2]{#2}
\providecommand{\BIBentrySTDinterwordspacing}{\spaceskip=0pt\relax}
\providecommand{\BIBentryALTinterwordstretchfactor}{4}
\providecommand{\BIBentryALTinterwordspacing}{\spaceskip=\fontdimen2\font plus
\BIBentryALTinterwordstretchfactor\fontdimen3\font minus \fontdimen4\font\relax}
\providecommand{\BIBforeignlanguage}[2]{{%
\expandafter\ifx\csname l@#1\endcsname\relax
\typeout{** WARNING: IEEEtran.bst: No hyphenation pattern has been}%
\typeout{** loaded for the language `#1'. Using the pattern for}%
\typeout{** the default language instead.}%
\else
\language=\csname l@#1\endcsname
\fi
#2}}
\providecommand{\BIBdecl}{\relax}
\BIBdecl

\bibitem{lakhotia2021generative}
K.~Lakhotia, E.~Kharitonov, W.-N. Hsu, Y.~Adi, A.~Polyak, B.~Bolte, T.-A. Nguyen, J.~Copet, A.~Baevski, A.~Mohamed \emph{et~al.}, ``On generative spoken language modeling from raw audio,'' \emph{Transactions of the Association for Computational Linguistics}, 2021.

\bibitem{chou2025flow}
J.-C. Chou, J.~Zhou, and K.~Livescu, ``{Flow-SLM}: joint learning of linguistic and acoustic information for spoken language modeling,'' in \emph{Proc. ASRU}, 2025.

\bibitem{sugiura2025llama}
I.~Sugiura, S.~Kurita, Y.~Oda, and R.~Higashinaka, ``{Llama-Mimi}: speech language models with interleaved semantic and acoustic tokens,'' \emph{arXiv preprint arXiv:2509.14882}, 2025.

\bibitem{arora2025landscape}
S.~Arora, K.-W. Chang, C.-M. Chien, Y.~Peng, H.~Wu, Y.~Adi, E.~Dupoux, H.-Y. Lee, K.~Livescu, and S.~Watanabe, ``On the landscape of spoken language models: a comprehensive survey,'' \emph{Transactions on Machine Learning Research}, 2025.

\bibitem{defossez2024moshi}
A.~D{\'e}fossez, L.~Mazar{\'e}, M.~Orsini, A.~Royer, P.~P{\'e}rez, H.~J{\'e}gou, E.~Grave, and N.~Zeghidour, ``{Moshi}: a speech-text foundation model for real-time dialogue,'' \emph{arXiv preprint arXiv:2410.00037}, 2024.

\bibitem{hu2025salm}
K.~Hu, E.~Hosseini-Asl, C.~Chen, E.~Casanova, S.~Ghosh, P.~{\.Z}elasko, Z.~Chen, J.~Li, J.~Balam, and B.~Ginsburg, ``{SALM-Duplex}: efficient and direct duplex modeling for speech-to-speech language model,'' \emph{Proc. Interspeech}, 2025.

\bibitem{nguyen2025spirit}
T.~A. Nguyen, B.~Muller, B.~Yu, M.~R. Costa-Jussa, M.~Elbayad, S.~Popuri, C.~Ropers, P.-A. Duquenne, R.~Algayres, R.~Mavlyutov \emph{et~al.}, ``{Spirit-LM}: interleaved spoken and written language model,'' \emph{Transactions of the Association for Computational Linguistics}, 2025.

\bibitem{maimon2025scaling}
G.~Maimon, M.~Hassid, A.~Roth, and Y.~Adi, ``Scaling analysis of interleaved speech-text language models,'' in \emph{Proc. COLM}, 2025.

\bibitem{fang2025llama}
Q.~Fang, S.~Guo, Y.~Zhou, Z.~Ma, S.~Zhang, and Y.~Feng, ``{LLaMA-Omni}: seamless speech interaction with large language models,'' in \emph{Proc. ICLR}, 2025.

\bibitem{ghosh2026audio}
A.~Goel, S.~Ghosh, J.~Kim, S.~Kumar, Z.~Kong, S.-G. Lee, C.-H.~H. Yang, R.~Duraiswami, D.~Manocha, R.~Valle, and B.~Catanzaro, ``{Audio Flamingo} 3: advancing audio intelligence with fully open large audio language models,'' in \emph{Advances in Neural Information Processing Systems}, 2025.

\bibitem{lu2024desta}
K.-H. Lu, Z.~Chen, S.-W. Fu, H.~Huang, B.~Ginsburg, Y.-C.~F. Wang, and H.-y. Lee, ``{DeSTA}: enhancing speech language models through descriptive speech-text alignment,'' in \emph{Proc. Interspeech}, 2024.

\bibitem{lu2026desta2}
K.-H. Lu, Z.~Chen, S.-W. Fu, C.-H.~H. Yang, S.-F. Huang, C.-K. Yang, C.-E. Yu, C.-W. Chen, W.-C. Chen, C.-y. Huang \emph{et~al.}, ``{DeSTA2.5-Audio}: toward general-purpose large audio language model with self-generated cross-modal alignment,'' \emph{IEEE/ACM Transactions on Audio, Speech, and Language Processing}, 2026.

\bibitem{cuervo2024scaling}
S.~Cuervo and R.~Marxer, ``Scaling properties of speech language models,'' in \emph{Proc. EMNLP}, 2024.

\bibitem{ramapuram2026scaling}
J.~Ramapuram, E.~G. Dhekane, A.~Shidani, D.~Busbridge, B.~Mazoure, Z.~Gu, R.~Webb, T.~Likhomanenko, and N.~Jaitly, ``Scaling properties of continuous diffusion spoken language models,'' \emph{arXiv preprint arXiv:2604.24416}, 2026.

\bibitem{borsos2023audiolm}
Z.~Borsos, R.~Marinier, D.~Vincent, E.~Kharitonov, O.~Pietquin, M.~Sharifi, D.~Roblek, O.~Teboul, D.~Grangier, M.~Tagliasacchi \emph{et~al.}, ``{AudioLM}: a language modeling approach to audio generation,'' \emph{IEEE/ACM Transactions on Audio, Speech, and Language Processing}, 2023.

\bibitem{rouard2025continuous}
S.~Rouard, M.~Orsini, A.~Roebel, N.~Zeghidour, and A.~D{\'e}fossez, ``Continuous audio language models,'' in \emph{Proc. ICLR}, 2026.

\bibitem{yang2025generative}
S.-W. Yang, B.~Kim, K.-P. Huang, Q.~Tang, H.~Phan, B.-R. Lu, H.~Sundar, S.~Ghosh, H.-y. Lee, C.-C. Kao, and C.~Wang, ``Generative audio language modeling with continuous-valued tokens and masked next-token prediction,'' in \emph{Proc. ICML}, 2025.

\bibitem{chou2023toward}
J.-C. Chou, C.-M. Chien, W.-N. Hsu, K.~Livescu, A.~Babu, A.~Conneau, A.~Baevski, and M.~Auli, ``Toward joint language modeling for speech units and text,'' in \emph{Findings of EMNLP}, 2023.

\bibitem{hsu2026anatomy}
M.-H. Hsu, X.~Zhang, X.~Tian, J.~Zhang, and Z.~Wu, ``Anatomy of the modality gap: dissecting the internal states of end-to-end speech {LLMs},'' \emph{arXiv preprint arXiv:2603.01502}, 2026.

\bibitem{wang2026speech}
H.~Wang, H.~Wang, Y.~Guo, Z.~Li, C.~Du, and K.~Yu, ``Why do speech language models fail to generate semantically coherent outputs? a modality evolving perspective,'' in \emph{Proc. ICASSP}, 2026.

\bibitem{hsu2021hubert}
W.-N. Hsu, B.~Bolte, Y.-H.~H. Tsai, K.~Lakhotia, R.~Salakhutdinov, and A.~Mohamed, ``{HuBERT}: Self-supervised speech representation learning by masked prediction of hidden units,'' \emph{IEEE/ACM Transactions on Audio, Speech, and Language Processing}, 2021.

\bibitem{lipman2022flow}
Y.~Lipman, R.~T. Chen, H.~Ben-Hamu, M.~Nickel, and M.~Le, ``Flow matching for generative modeling,'' in \emph{Proc. ICLR}, 2023.

\bibitem{li2026back}
T.~Li and K.~He, ``Back to basics: let denoising generative models denoise,'' in \emph{Proc. CVPR}, 2026.

\bibitem{le2023voicebox}
M.~Le, A.~Vyas, B.~Shi, B.~Karrer, L.~Sari, R.~Moritz, M.~Williamson, V.~Manohar, Y.~Adi, J.~Mahadeokar \emph{et~al.}, ``{Voicebox}: text-guided multilingual universal speech generation at scale,'' in \emph{Advances in Neural Information Processing Systems}, 2023.

\bibitem{chen2025f5}
Y.~Chen, Z.~Niu, Z.~Ma, K.~Deng, C.~Wang, J.~JianZhao, K.~Yu, and X.~Chen, ``{F5-TTS}: a fairytaler that fakes fluent and faithful speech with flow matching,'' in \emph{Proc. ACL}, 2025.

\bibitem{ma2024emotion2vec}
Z.~Ma, Z.~Zheng, J.~Ye, J.~Li, Z.~Gao, S.~Zhang, and X.~Chen, ``emotion2vec: self-supervised pre-training for speech emotion representation,'' in \emph{Findings of ACL}, 2024.

\bibitem{he2024emilia}
H.~He, Z.~Shang, C.~Wang, X.~Li, Y.~Gu, H.~Hua, L.~Liu, C.~Yang, J.~Li, P.~Shi \emph{et~al.}, ``{Emilia}: an extensive, multilingual, and diverse speech dataset for large-scale speech generation,'' in \emph{Proc. SLT}, 2024.

\bibitem{pratap2020mls}
V.~Pratap, Q.~Xu, A.~Sriram, G.~Synnaeve, and R.~Collobert, ``{MLS}: a large-scale multilingual dataset for speech research,'' in \emph{Proc. Interspeech}, 2020.

\bibitem{raffel2020exploring}
C.~Raffel, N.~Shazeer, A.~Roberts, K.~Lee, S.~Narang, M.~Matena, Y.~Zhou, W.~Li, and P.~J. Liu, ``Exploring the limits of transfer learning with a unified text-to-text transformer,'' \emph{Journal of Machine Learning Research}, 2020.

\bibitem{panayotov2015librispeech}
V.~Panayotov, G.~Chen, D.~Povey, and S.~Khudanpur, ``{LibriSpeech}: an {ASR} corpus based on public domain audio books,'' in \emph{Proc. ICASSP}, 2015.

\bibitem{yeh2025whisper}
S.-L. Yeh, Y.~Meng, and H.~Tang, ``{Whisper} has an internal word aligner,'' in \emph{Proc. ASRU}, 2025.

\bibitem{hassid2023textually}
M.~Hassid, T.~Remez, T.~A. Nguyen, I.~Gat, A.~Conneau, F.~Kreuk, J.~Copet, A.~Defossez, G.~Synnaeve, E.~Dupoux \emph{et~al.}, ``Textually pretrained speech language models,'' in \emph{Advances in Neural Information Processing Systems}, 2023.

\bibitem{mehta2024openelm}
S.~Mehta, M.~H. Sekhavat, Q.~Cao, M.~Horton, Y.~Jin, C.~Sun, I.~Mirzadeh, M.~Najibi, D.~Belenko, P.~Zatloukal \emph{et~al.}, ``{OpenELM}: an efficient language model family with open training and inference framework,'' \emph{Workshop on Efficient Systems for Foundation Models II @ ICML2024}, 2024.

\bibitem{hwang2023torchaudio}
J.~Hwang, M.~Hira, C.~Chen, X.~Zhang, Z.~Ni, G.~Sun, P.~Ma, R.~Huang, V.~Pratap, Y.~Zhang \emph{et~al.}, ``{TorchAudio} 2.1: advancing speech recognition, self-supervised learning, and audio processing components for {PyTorch},'' in \emph{Proc. ASRU}, 2023.

\bibitem{radford2023robust}
A.~Radford, J.~W. Kim, T.~Xu, G.~Brockman, C.~McLeavey, and I.~Sutskever, ``Robust speech recognition via large-scale weak supervision,'' in \emph{Proc. ICML}, 2023.

\bibitem{team2025olmo}
{OLMo Team}, ``{OLMo} 3,'' \emph{arXiv preprint arXiv:2512.13961}, 2025.

\bibitem{li2025powsm}
C.-J. Li, K.~Chang, S.~Bharadwaj, E.~Yeo, K.~Choi, J.~Zhu, D.~Mortensen, and S.~Watanabe, ``{POWSM}: a phonetic open {Whisper}-style speech foundation model,'' in \emph{Proc. ACL}, 2026.

\bibitem{saeki2022utmos}
T.~Saeki, D.~Xin, W.~Nakata, T.~Koriyama, S.~Takamichi, and H.~Saruwatari, ``{UTMOS}: {UTokyo-SaruLab} system for {VoiceMOS} challenge 2022,'' \emph{Proc. Interspeech}, 2022.

\bibitem{kaplan2020scaling}
J.~Kaplan, S.~McCandlish, T.~Henighan, T.~B. Brown, B.~Chess, R.~Child, S.~Gray, A.~Radford, J.~Wu, and D.~Amodei, ``Scaling laws for neural language models,'' \emph{arXiv preprint arXiv:2001.08361}, 2020.

\bibitem{tsai2026false}
Y.-S. Tsai, Y.-C. Lin, H.-C. Chou, T.-W. Hsu, Y.-M. Hsu, C.~W. Chen, S.~Narayanan, and H.-y. Lee, ``The false resonance: a critical examination of emotion embedding similarity for speech generation evaluation,'' \emph{arXiv preprint arXiv:2604.26347}, 2026.

\bibitem{liu2026ministral}
A.~H. Liu, K.~Khandelwal, S.~Subramanian, V.~Jouault, A.~Rastogi, A.~Sad{\'e}, A.~Jeffares, A.~Jiang, A.~Cahill, A.~Gavaudan \emph{et~al.}, ``Ministral 3,'' \emph{arXiv preprint arXiv:2601.08584}, 2026.

\bibitem{team2024gemma}
G.~Team, ``{Gemma}: Open models based on gemini research and technology,'' \emph{arXiv preprint arXiv:2403.08295}, 2024.

\bibitem{qwen2024qwen2}
Q.~Team, ``{Qwen2.5} technical report,'' \emph{arXiv preprint arXiv:2412.15115}, 2024.

\bibitem{grattafiori2024llama}
A.~Grattafiori, A.~Dubey, A.~Jauhri, A.~Pandey, A.~Kadian, A.~Al-Dahle, A.~Letman, A.~Mathur, A.~Schelten, A.~Vaughan \emph{et~al.}, ``The {Llama} 3 herd of models,'' \emph{arXiv preprint arXiv:2407.21783}, 2024.

\end{thebibliography}
